\newif
\ifconfver
\confvertrue        
\ifconfver
    \documentclass[10pt,journal,final,twoside]{IEEEtran}
\else
    \documentclass[11pt,draftcls,onecolumn,draftclsnofoot]{IEEEtran}
    \usepackage{setspace}
\fi
\usepackage{xspace,empheq,fancybox,amsmath,amssymb,graphicx,epstopdf,epsfig,syntonly,times,amsthm} \usepackage{psfrag,color,bm,array,cite}
\usepackage{url,cite,footnote,xspace,syntonly,algorithm,algorithmic}
\usepackage{verbatim,multirow}
\usepackage[T1]{fontenc}
\usepackage{mathtools}
\usepackage{amsmath,amsfonts,amsthm,bm} 
\usepackage{graphicx}
\usepackage{mwe}
\usepackage{caption}
\usepackage{cite}
\usepackage{amsmath,amssymb,amsfonts}
\usepackage{algorithmic}
\usepackage{graphicx}
\usepackage{textcomp}
\usepackage{xcolor}
\usepackage{multicol, blindtext}
\usepackage{xspace,empheq,fancybox,amsmath,amssymb,graphicx,epstopdf,epsfig,syntonly,times,amsthm} \usepackage{psfrag,color,bm,array,cite}
\usepackage{url,cite,footnote,xspace,syntonly,algorithm,algorithmic}
\usepackage{verbatim,multirow}
\usepackage[T1]{fontenc}
\usepackage{mathtools}
\usepackage{subfig}
\usepackage{amsmath,amsfonts,amsthm,bm}
\usepackage{graphicx}
\usepackage{mwe}
\usepackage{caption}
\usepackage{stfloats}
\usepackage{amsfonts}
\usepackage{stackengine}
\usepackage{lipsum}
\usepackage{bbm}

\usepackage{xcolor}
\usepackage{array}
\usepackage{mathtools}
\makeatletter
\newcommand{\removelatexerror}{\let\@latex@error\@gobble}
\makeatother

\newcolumntype{P}[1]{>{\centering\arraybackslash}p{#1}}

\newtheorem{remark}{\bfseries Remark}
\input{mysymbol.sty}

\definecolor{orange}{rgb}{1,0.5,0}

\definecolor{color_yuqi}{RGB}{0, 0, 255}

\definecolor{color_yuqi2}{RGB}{155 225 152}

\allowdisplaybreaks

\usepackage{amsthm}

\DeclareMathOperator*{\argmin}{arg\,min}

\theoremstyle{remark}

\renewcommand{\parallel}{\mathrel{/\mkern-5mu/}}
\makeatletter
\newcommand{\notparallel}{%
  \mathrel{\mathpalette\not@parallel\relax}%
}
\newcommand{\not@parallel}[2]{%
  \ooalign{\reflectbox{$\m@th#1\smallsetminus$}\cr\hfil$\m@th#1\parallel$\cr}%
}
\makeatother

\begin{document}
\title{Machine Learning for Scalable and Optimal \\Load Shedding Under Power System Contingency} 

\author{
\IEEEauthorblockN{Yuqi Zhou},~\IEEEmembership{Member, IEEE},  and \IEEEauthorblockN{Hao Zhu},~\IEEEmembership{Senior Member, IEEE}

\thanks{\protect\rule{0pt}{3mm} This work has been supported by NSF Grants 2130706 and 2150571.
This work was partly authored by the National Renewable Energy Laboratory, operated by Alliance for Sustainable Energy, LLC, for the U.S. Department of Energy (DOE). The views expressed in the article do not necessarily represent the views of the DOE or the U.S. Government. The U.S. Government retains, and the publisher, by accepting the article for publication, acknowledges that the U.S. Government retains a nonexclusive, paid-up, irrevocable, worldwide license to publish or reproduce the published form of this work or allow others to do so for U.S. Government purposes. 

Y. Zhou is with the Power Systems Engineering Center, National Renewable Energy Laboratory, Golden, CO, 80401, USA (email: {yuqi.zhou{@}nrel.gov}).

H. Zhu is with the Chandra Family Department of Electrical \& Computer Engineering, The University of Texas at Austin, Austin, TX, 78712, USA (email: haozhu{@}utexas.edu).
}
}

\renewcommand{\thepage}{}
\maketitle
\pagenumbering{arabic}

\begin{abstract}
Prompt and effective corrective actions in response to unexpected contingencies are crucial for improving power system resilience and preventing cascading blackouts. The optimal load shedding (OLS) accounting for network limits has the potential to address the diverse system-wide impacts of contingency scenarios as compared to traditional local schemes. However, due to the fast cascading propagation of initial contingencies, real-time OLS solutions are challenging to attain in large systems with high computation and communication needs. In this paper, we propose a decentralized design that leverages offline training of a neural network (NN) model for individual load centers to autonomously construct the OLS solutions from locally available measurements. Our learning-for-OLS approach can greatly reduce the computation and communication needs during online emergency responses, thus preventing the cascading propagation of contingencies for enhanced power grid resilience. Numerical studies on both the IEEE 118-bus system and a synthetic Texas 2000-bus system have demonstrated the efficiency and effectiveness of our scalable OLS learning design for timely power system emergency operations.

\end{abstract}


\section{Introduction}
\label{sec:intro}

Prompt and effective responses to emergency events are crucial for enhancing power system resilience. Cascading failures could be mitigated via corrective actions by restoring power balance and maintaining network limits. Load shedding is such an emergency response commonly employed by grid operators via disconnecting the non-critical loads. Unlike normal operations, the decision-making of emergency responses is timing-critical. Due to the computation and communication needs for effective load shedding, machine learning (ML) approaches provide a unique opportunity for designing timely emergency responses by leveraging their online prediction capability.

Traditional load shedding approaches rely on localized automatic control mechanisms. To quickly restore \textit{power balance}, Under Frequency Load Shedding (UFLS) \cite{omar2010under,ketabi2014underfrequency,
larik2018improved} and  Under Voltage Load Shedding (UVLS) \cite{larik2019critical,mozina2007undervoltage,amraee2007enhanced} are designed to respond to the drops in locally observed frequency or voltage using a fixed rate of reduction; see also the NERC standards in \cite{NERC}. This proportional rule is easy to implement locally but fails to account for the grid-wide heterogeneous effects of the contingency scenario such as severe congestion on certain lines. To address this issue, the optimization-based load shedding scheme  (e.g., \cite{coffrin2018relaxations,rhodes2020balancing,lin2016admm}) 
has become popular, relying on centralized dispatch using advanced monitoring and communications across the system. The optimal load shedding (OLS) can effectively mitigate the risks of cascading failures through strategically targeting system-wide congestion or \textit{network limits}. For example, convex relaxation is used in \cite{coffrin2018relaxations,fernandes2008load} to address the nonlinearity of AC power flow, while the fairness of power shut-offs is considered in \cite{kody2022sharing}. 
Nevertheless, solving this optimization in real time is still challenged by the high computational complexity and communication needs.

Similar to the rising adoption of machine learning (ML) for optimal power flow (OPF) \cite{chen_opf2023}, there is a surge of interest in using ML for online AC-OLS implementation. 
In a stressed system with power imbalance, failure to perform effective load shedding to stabilize the grid could trigger cascading failures within seconds to a few minutes \cite{pahwa2013load,sinha2019optimal,song2015dynamic}. Thus, to fully leverage the benefits of a centralized load shedding scheme while addressing the time-sensitive nature of such implementation, ML is ideally suited to enable prompt decision-making. A graph neural network-based approach is developed in \cite{kim2019graph} that requires the system-wide contingency information and observability, while reinforcement learning is considered in \cite{vu2021safe,huang2019adaptive,zhang2024multi} for obtaining dynamic load shedding policies using the grid-wide system states as the input. Please see Table \ref{tab:LR} for a detailed comparison of existing approaches.
Although a centralized optimization/ML is suitable for grid decision-making during normal conditions, it is less attractive for timing-critical emergency responses due to the stringent requirements on sensing and communications capabilities. 
Therefore, it remains to develop autonomous OLS solutions that can cope up with the real-time needs at distributed load centers by providing timely corrective actions in response to sudden contingencies.

This paper aims to develop a learning-for-OLS framework to enable individual load centers to promptly form their own OLS decisions in a decentralized fashion. We first formulate the  AC-OLS problem of determining the amount of nodal load shedding, to restore the system-wide power balance, and most importantly, mitigate the violations of network limits. Upon solving this problem under a range of loading conditions and contingency scenarios, one can learn the decision rules that map from input system conditions to target OLS outputs through offline training of NN models to capitalize on their excellent approximation capability. 
One notable feature is the \textit{decentralized} design of the proposed NN-based OLS decision rules, which have been constructed from the \textit{locally available measurements} at each load center only. Different from our preliminary work \cite{zhou2022scalable}, this paper has judiciously designed the NN prediction outputs to be the nodal Lagrange multipliers of the OLS problem. The latter is known to enjoy a local dependence property \cite{jia2013impact,liu2022topology}, thus ideally suited for the decentralized design. We have further developed the identifiability analysis based on the simplified DC power flow for understanding the inference capability of using local data only.  
The scalability of our design can significantly reduce the offline training time as compared to a centralized framework. Most importantly, these resultant local OLS decision rules allow for each load center to promptly and autonomously react to the specific contingency in real time, without the need to communicate to the control center. This way, it effectively eliminates the negative impacts of communication latency and reliability issues that are critical for emergency operations. The main contributions of our work include the following:
\begin{itemize}
  \item A decentralized learning-for-OLS framework is proposed, allowing each load center to promptly predict its own OLS solution in response to emergency events.
  \item The ML algorithm is designed to use local measurements to quickly predict local dual variables of the OLS problem, enabling accurate mapping to corrective actions during real-time operations.
  \item An identifiability analysis is formally performed to discuss the use of solely local measurements and support the validity of our decentralized OLS design.
\end{itemize}

This paper is organized as follows. Section \ref{sec:ps} formulates the AC-OLS as a nonlinear optimization problem. Section \ref{sec:learning} presents the design of our proposed scalable OLS method, building upon offline training and computation to attain autonomous online responses. The identifiability analysis is presented in Section \ref{sec:analysis}, to showcase the capability of differentiating contingencies using local data. In Section \ref{sec:sr}, numerical simulations on both the IEEE 118-bus case and the synthetic Texas 2000-bus system have demonstrated the accuracy and efficiency of the scalable OLS design for real-time emergency operations. Last, the paper is concluded in Section \ref{sec:con}.

\begin{table*}[t!]
\caption{Comparisons of Existing Approaches and Our Work}
\label{tab:LR}
\centering
\begin{tabular}{|p{2cm}||p{2.2cm}|p{2.4cm}|p{3cm}|p{4.5cm}|}
\hline
{} & {Objective} & {Method} & {Decentralization} & {Techniques} \\
\hline
{\cite{larik2018improved}} & {Power balance} & {Optimization} & {Centralized} & {Particle swarm optimization} \\
{\cite{coffrin2018relaxations,fernandes2008load}} & {Network limits} & {Optimization} & {Centralized} & {Convex relaxation} \\
{\cite{rhodes2020balancing}} & {Network limits} & {Optimization} & {Centralized} & {Deterministic optimization} \\
{\cite{awad2014optimal,moradi2010optimal}} & {Network limits} & {Optimization} & {Centralized} & {Genetic algorithm} \\
{\cite{luo2024optimal,gu2014adaptive,xu2011stable}} & {Power balance} & {Multi-agent system} & {Distributed} & {Consensus-based algorithm} \\
{\cite{kim2019graph}} & {Network limits} & {Machine learning} & {Centralized} & {Graph neural networks} \\
{\cite{zhu2017load,yang2018pmu}} & {Power balance} & {Machine learning} & {Centralized} & {Support vector machine} \\
{\cite{voumvoulakis2006decision,zhu2019response}} & {Power balance} & {Machine learning} & {Centralized} & {Decision tree} \\
{\cite{vu2021safe,huang2022learning,zhang2024multi}} & {Power balance} & {Machine learning} & {Centralized} & {Reinforcement learning} \\
\hline
{Our work} & {Network limits} & {Machine Learning} & {Decentralized} & {NN for a decentralized design}\\
\hline
\end{tabular}
\end{table*}

\vspace*{5pt}

\textit{Notation:} $(\cdot)^{\mathsf T}$ stands for transposition; $(\cdot)^{\star}$ stands for the conjugate of a complex number; $(\cdot)^{*}$ stands for the optimal solution; $|\cdot|$ denotes the absolute value of a complex number; $\|\cdot\|$ denotes the vector Euclidean norm; $\mathbb{R}$ denotes real numbers; $\mathbb{R}^{+}$ denotes positive real numbers.

\section{AC-OLS Problem Formulation} \label{sec:ps}

We formulate the optimal load shedding (OLS) problem under the nonlinear AC power flow model. Consider a transmission system with $N$ buses and $L$ lines collected in the sets $\cal N$ and $\cal L$, respectively. Let $\bm{Y}= \bm{G} + \mathrm{j} \bm{B}$ denote the complex network admittance matrix of size $N\times N$. Per bus $i$, we denote the complex power output from its connected generation by $p_{i}^{g}+\mathrm{j} q_{i}^{g}$, and from load demand by $p_{i}^{d}+\mathrm{j}q_{i}^{d}$. In addition, let $V_{i} \angle \theta_{i}$ denote the bus voltage phasor with the magnitude $V_{i}$ and phase angle $\theta_{i}$. Per line $(i,j) \in \cal L$, the complex power flow from bus $i$ to $j$ becomes:
\begin{align} 
    S_{ij} = {Y}_{ij}^{\star} V_{i} ^2 - {Y}_{ij}^{\star} V_{i} V_{j} \angle(\theta_i-\theta_j) \label{eq:acpf}
\end{align}
using the $(i,j)$-th entry of matrix $\bm{Y}$.

\begin{figure}[t!]
\centering
\includegraphics[trim=0cm 0cm 0cm 0cm,clip=true,width=0.9\linewidth]{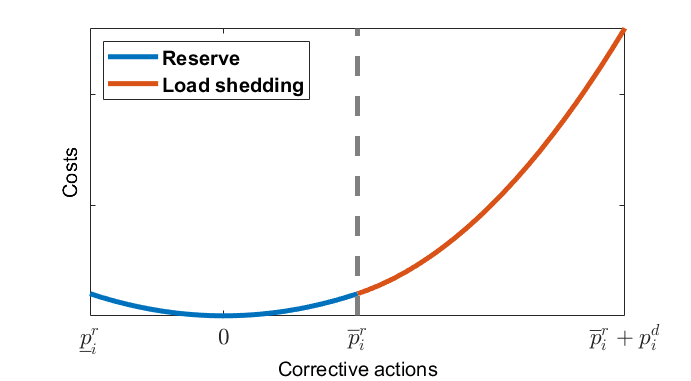}
\caption{A piece-wise quadratic cost function for the active power flexibility (both reserve and load shedding) per bus $i$.}
\vspace*{-2mm}
\label{fig:cost_function}
\end{figure}

The AC-OLS problem is essentially an extension of the optimal power flow (OPF). The latter determines the optimal set-points of controllable resources under \textit{normal operations} by accounting for the grid operation limits. During \textit{emergency operations} arising from large contingencies, AC-OPF can become infeasible with insufficient resource flexibility or transfer capability. 
Thus, using generation reserves, and more drastic corrective actions like load shedding, are imperative for restoring power balance while satisfying operational limits. With known nominal injection $p_{i} := p_{i}^{g} - p_{i}^{d}$ and similarly $q_{i}$ per bus $i$, the OLS problem aims to determine the amount of generation reserves and the load demand to shed, as
\begin{subequations} \label{eq:OLS}
\begin{align}
\min \; \, \, & \textstyle \sum_{i=1}^{N} \, c_{i}^{p}(p_{i}^{s})   \label{eq:OLS_a}\\
\textrm{s.~t.} \; \, \, &  V_{i} \in \mathbb{R}^{+}, \theta_{i} \in \mathbb{R}, 
  p_{i}^{s} \in \mathbb{R}, q_{i}^{s} \in \mathbb{R}, \, \forall i \in \cal N   \label{eq:OLS_c}\\
  & {\underline{V}_{i}} \leq V_{i} \leq {\overline{V}_{i}}, \: \: {\underline{\theta}_{i}} \leq \theta_{i} \leq {\overline{\theta}_{i}} \label{eq:OLS_d}\\
  & \underline{p}_{i}^{r} \leq p_{i}^{s} \leq \overline{p}_{i}^{r} + p_{i}^{d}, \: \: \underline{q}_{i}^{r} \leq q_{i}^{s} \leq \overline{q}_{i}^{r} + q_{i}^{d} \label{eq:OLS_e}\\
  & |S_{ij}| \leq \overline{S}_{ij},\, \forall (i,j) \in \cal L   \label{eq:OLS_f}\\
  & \theta_{ij} = \theta_{i} - \theta_{j}, \, \forall (i,j) \in \cal L \label{eq:OLS_g}\\
  & p_{i} + p_{i}^{s} = \textstyle \sum_{j=1}^{N}  V_{i} V_{j} \left(G_{ij}\cos \theta_{ij}  + B_{ij}\sin \theta_{ij}\right)\label{eq:OLS_h} \\
  & q_{i} + q_{i}^{s} =  \textstyle \sum_{j=1}^{N} V_{i}V_{j} \left(G_{ij}\sin \theta_{ij}  - B_{ij}\cos \theta_{ij}\right). \label{eq:OLS_i}
\end{align}
\end{subequations}
For the decision variables given in \eqref{eq:OLS_c}, constraints in \eqref{eq:OLS_d}-\eqref{eq:OLS_f} enforce the operational limits or resource budget. Note that the line thermal limit in \eqref{eq:OLS_f} in terms of apparent power flow can also include other metrics such as line current or active power flow. Finally, the equality constraints in \eqref{eq:OLS_g}-\eqref{eq:OLS_i} represent the AC power flow model in  \eqref{eq:acpf}. 
Note that our AC-OLS formulation includes both active and reactive powers as decision variables, with the cost function \eqref{eq:OLS_a} focusing on active power only.
Similar to AC-OPF, the balance of reactive power in \eqref{eq:OLS} is crucial for maintaining voltage stability, and it could automatically adjusted via local voltage control in practice. Thus, while \eqref{eq:OLS} follows from earlier work \cite{coffrin2018relaxations,larik2018improved,fernandes2008load} to include reactive power flexibility, the proposed method does not need to consider the prediction of these decisions.
A simplified DC-OLS formulation with active power decisions only is also possible; see e.g., \cite{xu2001optimal,lin2016admm}. Note that most of the cases, the AC-OLS problem \eqref{eq:OLS} will automatically search for an optimal operating point that ensures feasibility, thanks to the flexibility provided by load shedding and reserves. However, in very extreme cases, it is possible that the contingency can cause infeasibility of the problem, and such scenarios can be handled by further allowing for disconnection of other components \cite{coffrin2018relaxations} in the network.

\begin{remark}[Cost function design]
The decision variables $\{p_{i}^{s}\}$ in \eqref{eq:OLS} represent the flexibility from both utilizing generation reserves and performing load shedding at each bus. As shown in Fig.~\ref{fig:cost_function}, 
we consider a piece-wise quadratic cost:
\begin{align}  \label{eq:cost}
c_{i}^{p}(p_i^s)\!: =\!
    \begin{cases}
        a_{i,1}(p_{i}^{s})^{2} , & \!\! \! \underline{p}_{i}^{r} \leq p_{i}^{s} \leq \overline{p}_{i}^{r} \\
         a_{i,2}(p_{i}^{s})^{2} + b_{i,2} {p}_{i}^{s}+ c_{i,2},\!\! &  \!\! \!\overline{p}_{i}^{r} \leq p_{i}^{s} \leq \overline{p}_{i}^{r} + \overline{p}_{i}^{d}
    \end{cases}
\end{align}
Here, $\underline{p}_{i}^{r}$ and $\overline{p}_{i}^{r}$ 
represent the limits of downward and upward reserves, respectively. Once all the reserves are utilized, the more expensive load-shedding option can be employed, capped at a maximum of $\overline{p}_{i}^{d}$. The latter can be smaller than the full loading of $p_i^d$, due to the presence of non-flexible critical loads. The two quadratic functions vary in incremental costs, with load shedding being more costly to initiate, and thus reserves are prioritized over load shedding.
\end{remark}

Under such cost function design, the optimization problem \eqref{eq:OLS} will automatically determine the use of reserves and load shedding. Specifically, when no correction actions are needed, the cost function is minimized at $p^{s}_{i} = 0$. When minor corrections such as reserves are necessary, the variable $p^{s}_{i}$ takes values in the range of $[\underline{p}_{i}^{r}, 0]$ (downward reserves) or $[0, \overline{p}_{i}^{r}]$ (upward reserves). Whenever the optimal solution falls within $[\underline{p}_{i}^{r}, \overline{p}_{i}^{r}]$ (blue piece in Fig. \ref{fig:cost_function}), it indicates that only reserves are utilized. When all the reserves are utilized, the more expensive load shedding operations are activated. Solutions $p^{s*}_{i}$ in this scenario (orange piece in Fig. \ref{fig:cost_function}) indicate that both reserves and load shedding are utilized, where the reserves are equal to $\overline{p}_{i}^{r}$ and the optimal load shedding equals $p^{s*}_{i} - \overline{p}_{i}^{r}$.

In addition, the coefficient $a_2$ can vary among the buses based on the priority of the connected loads. 
For the function \eqref{eq:cost}, the following conditions hold at the point $\overline{p}_{i}^{r}$ in general:
\begin{align}
    a_{i,1}(\overline{p}_{i}^{r})^{2} &=  a_{i,2}(\overline{p}_{i}^{r})^{2} + b_{i,2}\overline{p}_{i}^{r} + c_{i,2}, \label{eq:con_1}\\
     2a_{i,1}\overline{p}_{i}^{r} &\leq  2a_{i,2}\overline{p}_{i}^{r} + b_{i,2}. \label{eq:con_2} 
\end{align}
The first condition guarantees the continuity of $c_i^p$, while the second ensures the incremental cost of reserves will not exceed that of load shedding. Hence, the AC-OLS problem mainly differs from AC-OPF in terms of the level of flexibility that each load center can provide. It is also worth pointing out that while our OLS problem \eqref{eq:OLS} only considers load shedding and regulating reserves, other corrective actions such as topology optimization and full de-energization of system components can be included, as well; see e.g., \cite{coffrin2018relaxations,rhodes2020balancing}. Notably, in this work, we focus on optimization models that allow for rapid adjustments to the system’s operating point to prevent power imbalances. After the load shedding actions are initiated, the dynamic behavior of the system will be managed with frequency control \cite{machowski2020power,bevrani2014robust}. The process generally involves primary control, secondary control (e.g., Automatic Generation Control), and tertiary control. These mechanisms are essential for managing frequency deviations \cite{garcia2021requirements} and thus ensure system stability.

\begin{remark}[Solving AC-OLS]
As a nonlinear program (NLP), the AC-OLS problem is nonconvex and generally NP-hard, similar to the AC-OPF \cite{castillo2013survey}. Some convex relaxation methods can be adopted to tackle the nonconvexity of AC power flow equations, see the OPF work in \cite{low2014convex,kocuk2016strong} and similarly the OLS one \cite{coffrin2018relaxations}. Fortunately, open-source packages such as MATPOWER and JuMP can be used to solve the NLP form of AC-OLS for practically sized power systems. 
\end{remark}

\section{Scalable OLS Design} \label{sec:learning}

Albeit solvable in a centralized fashion, real-time OLS decision-making can significantly benefit from a decentralized learning approach to accelerate the responses to emergency events. 
During the onset of contingency, corrective actions are pressingly needed in order to mitigate severe damages and the chances of cascading failures to the rest of the grids. Nonetheless, solving the centralized OLS problem in a timely manner could be impractical, due to its needs of communications and network coordination. In particular, an effective online implementation of centralized OLS critically requires high-rate low-latency communications, such that the control center can promptly acquire system-wide information and further dispatch decentralized resources.  However, the availability of such communication capability is challenging, especially during weather disasters or other disruptive events. 
Hence, to enable fast and effective OLS actions in real time, we put forth a scalable learning framework by leveraging offline training to facilitate prompt online decision-making.

Our proposed design aims to enable each load center to predict its own OLS decisions from locally available measurements only. 
For the prediction output, we choose the local \textit{marginal costs} of AC-OLS, namely the sensitivity factors of the overall cost over each individual nodal flexibility. This is related to the Lagrange multipliers-based duality analysis for the active and reactive power balance constraints in \eqref{eq:OLS_g}-\eqref{eq:OLS_i}. As shown by the recent OPF learning work \cite{chen2022learning,liu2022topology,singh2021learning}, predicting the locational marginal prices (LMPs), as the sensitivity counterpart for the OPF problem, can be much more effectively done using localized information only. This is because the LMPs enjoy the \textit{locality} property \cite{jia2013impact}; and similarly for the sensitivity analysis for generalized grid optimization problems like OLS. Inspired by these OPF-learning approaches, our decentralized OLS learning design will similarly predict the local marginal costs instead of the primal variables.

Let ${{\alpha}_i}$ represent the Lagrange multipliers for constraint \eqref{eq:OLS_h} per bus $i$. The optimality conditions by taking the upper/lower limits in \eqref{eq:OLS_e} as implicit constraints, state that the optimal solutions for ${p}^{s}_{i}$ become:
\begin{align}
    {{\hat{p}}^{s}_{i}} & =  \argmin_{ \underline{p}_{i}^{r} \leq p_{i}^{s} \leq \overline{p}_{i}^{r} + p_{i}^{d} } \: {c^{p}_i}({p^{s}_i}) - {{{\alpha}_i}} {p}^{s}_{i}, \label{eq:op_1}
\end{align}
The relation is in line with the LMP-based OPF duality analysis, where the prices fully determine the generator dispatch. While the duality analysis of AC-OLS or AC-OPF is an important topic, it goes beyond the scope of our work; see e.g.,  earlier work   \cite{conejo2005locational,li2007dcopf,garcia2020generalized} for more details. Using the piece-wise quadratic cost function in \eqref{eq:cost}, we can express the unique optimal solutions in closed-form:

\begin{align} \label{eq:close_form}
{\hat{p}}^{s}_{i} =
    \begin{cases}
        \left(\frac{{\alpha}_i - b_{i,2}}{2a_{i,2}}\right)_{[ \overline{p}_{i}^{r},  \overline{p}_{i}^{r}+ p_i^d]}, & \text{if }   {\alpha}_i > 2a_{i,2} \overline{p}_{i}^{r} + b_{i,2} \\
        \left(\frac{{\alpha}_i}{2a_{i,1}}\right)_{[\underline{p}_{i}^{r},  \overline{p}_{i}^{r} ]}, & \text{if }   
        {\alpha}_i \leq 2a_{i,2}\overline{p}_{i}^{r}+b_{i,2}\\
    \end{cases}
\end{align}
where the subscript indicates the projection to the corresponding box interval. 
Upon predicting the marginal costs, namely ${{\hat{\alpha}}_{i}}$, the load center at bus $i$ can uniquely determine its own load shedding decisions ${\hat{p}}^{s}_{i}$. This equivalent condition will be utilized in our OLS learning design.

\begin{figure}[t!]
\centering
\includegraphics[trim=0cm 0cm 0cm 0.3cm,clip=true,width=0.9\linewidth]{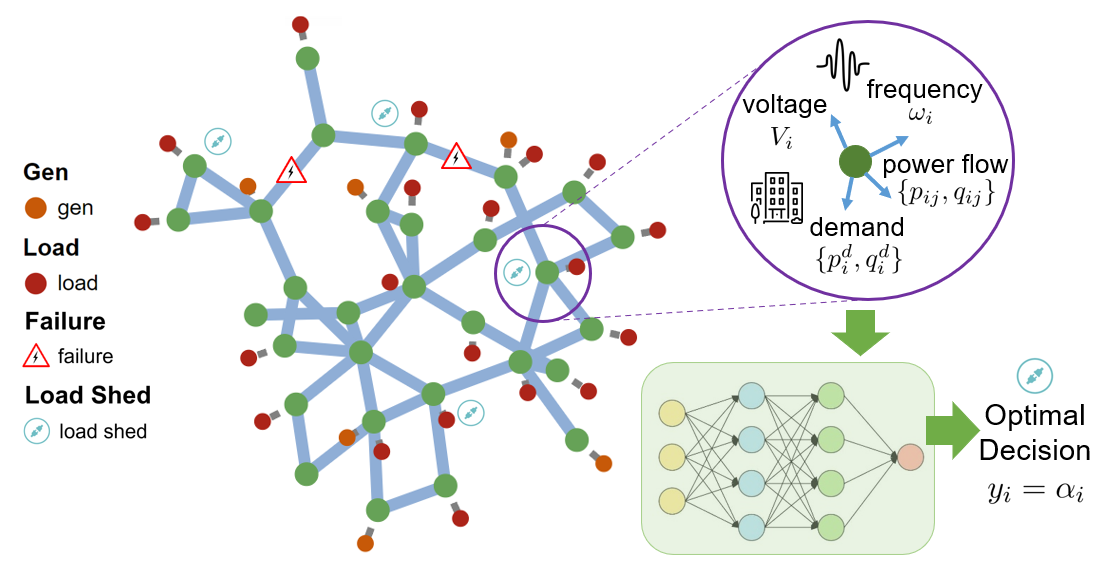}
\caption{Illustration of the proposed scalable OLS design.}
\vspace*{-2mm}
\label{fig:structure}
\end{figure}

Fig.~\ref{fig:structure} illustrates the overall architecture of our scalable OLS design using an example system of the IEEE 30-bus test case.
At each load center (or possibly a sub-area), local measurements such as voltage, frequency, power demand, and incident line flows are readily available from localized meters. Using these data during both pre- and post-contingency, the load center aims to predict the aforementioned OLS outputs, the marginal costs, and accordingly, optimal decisions, based upon a pre-trained neural network (NN). This way, no real-time communications are required between the control center and distributed loads. Of course, the feasibility of such a direct prediction from limited data within a local area comes into question. While there are no communications to exchange information, the post-contingency data at each load center reflect the physics of the underlying interconnection and contingency scenario according to the network power flow coupling. Thus, if each potential contingency exhibits a unique pattern of power flow changes at any load center, it becomes possible to infer the associated OLS decisions from its subset of measurements. We will formally analyze this inference capability in Section \ref{sec:analysis}.

To enable the scalable OLS design, extensive offline computations are utilized to obtain the best mapping from the available measurements to real-time OLS decisions. We consider the standard feedforward NN for this scalable mapping, as the input-output variables are localized within each load center $i$.
To formalize this prediction, we consider the input feature vector collecting all possible local measurements, given by:
\begin{align} \label{eq:input}
  \bm{x}_{i} = \left[p_{i}^{d}, q_{i}^{d}, V_{i}, \omega_{i}, \{p_{ij}\}_{j\in\ccalN_i}, \{q_{ij}\}_{j\in\ccalN_i}   \right],~\forall i,
\end{align}
which includes local active/reactive demands, voltage magnitude, electric frequency, and incident active/reactive line flows. Note that both pre- and post-contingency values are used, except for the demands which could be unchanged throughout the contingency. Using both conditions increases the possibility of inferring the underlying contingency and thus the associated OLS decisions, as the changes from pre- to post-contingency power flows can differentiate the specific contingency. For example, both $V_{i}$ and $\omega_{i}$ are good indicators of the system loading stress level.
Furthermore, the changes in line flows can effectively provide the identifiability on the location of outaged lines due to the power flow redistribution. As discussed earlier, we consider the NN outputs to be the local marginal costs of the AC-OLS problem, as given by:
\begin{align} \label{eq:output}
  {y}_{i} = \alpha_{i}.
\end{align}
This output per bus $i$ enables us to directly compute the optimal corrective actions by comparing to the local OLS costs as in \eqref{eq:op_1}. 
Convenient implementation of this step is possible during the online phase based upon \eqref{eq:close_form}, thus attaining a fully decentralized OLS decision-making as long as the load center at bus $i$ can predict ${y}_{i}$.

We can obtain the decision rule that maps from local data $\bm{x}_{i}$ to local outputs ${y}_{i}$, namely, $f_i(\bm{x}_{i}; \bm{\varphi}_i) \rightarrow {y}_{i}$ with $\bm{\varphi}_i$ denoting the NN parameters to be trained. To pursue a powerful decision rule $f_i(\cdot; \bm{\varphi}_i)$, massive data samples will be generated during the offline training to cover the possible operating conditions and contingency scenarios.
Thus, it is beneficial to randomize the system loading with a sufficiently large range, as well as to consider all possible contingency scenarios that are likely to occur in the upcoming operating horizon (next hour or day). Upon drawing samples of system loading and contingency scenarios, the offline computation generates the corresponding input-output pairs using model-based simulations. Specifically,  the input $\bm{x}_i$ can be computed by power system simulators such as MATPOWER through steady-state power flow analysis, except for the frequency input $\omega_i$. The latter could be approximated instead, using the difference between the total generated powers during  pre- and post-contingency; see e.g., \cite[Ch.~12]{glover2012power}. 
Meanwhile, the output ${y}_{i}$ can be obtained using grid optimization solvers based on MATPOWER. Basically, the AC-OLS problem \eqref{eq:OLS} can be solved to find the dual variables in ${y}_{i}$ based on the loading and contingency information.
Upon completing the training of bus $i$'s OLS rule using the samples of $\{\bm{x}_i,{y}_{i}\}$, the updated NN model will be sent to the load center on a regular basis (from hourly to daily), enabling a fully decentralized online implementation.  
In a nutshell, the proposed scalable OLS design leverages the extensive computation capability at the control center by performing offline training of OLS decision rules in order to empower the online implementation of decentralized OLS corrective actions.

\begin{figure}[t!] 
  \centering 
  \begin{minipage}{0.48\textwidth} 
  \begin{algorithm}[H]
  \caption{{Decentralized Learning for Scalable OLS}} 
  \label{alg:alg1}
  \begin{algorithmic}[1]
  \renewcommand{\algorithmicrequire}{\textbf{{Input:}}}
  \renewcommand{\algorithmicensure}{\textbf{{Output:}}}
  \REQUIRE {$\{C_{j}\}, \forall j = 1, \cdots, J; \: \{\bm{p}^{d}_{k}\}, \forall k = 1, \cdots, K$}
  \ENSURE {$\hat{p}^{s}_{i}, \forall i = 1, \cdots, N$}
  \vspace{1mm}
  \\ \textbf{\textit{{Collect optimization results (offline)}}} 
  \FOR {$j = 1$ to $J$}
  \FOR {$k = 1$ to $K$}
  \STATE {Solve optimization \eqref{eq:OLS}, and collect training samples $\bm{x}_{i} = \left[p_{i}^{d}, q_{i}^{d}, V_{i}, \omega_{i}, \{p_{ij}\}, \{q_{ij}\}   \right], \bm{y}_{i} = \alpha_{i}, \forall i$.}
  \ENDFOR
  \ENDFOR
  \vspace{1mm}

  \textbf{\textit{{Training the ML model (offline)}}} 
  \FOR {$i = 1$ to $N$}
  \STATE {Training for each load center to obtain the local decision rule $f_{i} (\cdot; \varphi_{i})$.}
  \ENDFOR

  \vspace{1mm}

  \textbf{\textit{{Real-time load shedding prediction (online)}}}
  \STATE {Each load center $i$ collects its real-time local measurements $\hat{\bm{x}}_{i}$.}
  \STATE {Calculate the predicted dual variable $\hat{\alpha}_{i} = f_{i} (\hat{\bm{x}}_{i}; \varphi_{i})$.}
  \STATE {Obtain the optimal corrective actions $\hat{p}^{s}_{i}$ via \eqref{eq:close_form}.}
  \RETURN {$\hat{p}^{s}_{i}, \forall i = 1, \cdots, N$}
  \end{algorithmic}
  \end{algorithm}
  \end{minipage}
\end{figure}

For the parameter $\bm{\varphi}_i$ per load center $i$, the feedforward NN model consists of multiple hidden layers between the input and output. With the first layer  $\bm {z}_i ^0 = \bm x_i$ incorporating the input feature, each layer $k$ can be represented as:
\begin{align} 
\bm{z}^{k}_{i} = \sigma \left(\bm{W}^{k} \bm{z}^{k-1}_{i} + \bm{b}^{k} \right),~~\forall k = 1,\ldots, K
\end{align}
where the final layer $\bm{z}^{K}_{i} \rightarrow y_i$ predicts the output target. Basically, the NN parameter vector $\bm{\varphi}_i$ includes the weight matrices $\{\bm{W}^{k}\}$ and bias vectors $\{\bm{b}^{k}\}$ for every layer $k$. There is also a nonlinear activation function  $\sigma (\cdot)$, used to achieve an expressive functional mapping that goes beyond linearity. Common choices of $\sigma (\cdot)$ include sigmoid and ReLU. 
To determine $\bm{\varphi}_i$, we use the mean squared error (MSE) metric as the loss function to minimize, which can be easily trained  by popular NN solvers like PyTorch. The detailed implementation of the scalable OLS algorithm is summarized in Algorithm \ref{alg:alg1}. The whole process starts with generating a prioritized list of contingencies $\{C\}$ and a sufficient number of load samples $\{\bm{p}^{d}\}$. Following this, the OLS problem is solved repeatedly to generate training samples. The training is conducted individually for each load center to obtain local decision rules. During real-time operations, each load center uses local measurements to predict its own dual variable ${\hat\alpha}_{i}$, then determines corrective actions through optimality condition \eqref{eq:close_form}. Note that both optimization solving and ML training are conducted offline, which enables accurate prediction of OLS decisions for each load center in real time.

\begin{remark}[Safety of OLS learning]
Corrective actions such as load shedding need to be effective and safe during online implementation. Such considerations can be incorporated into the offline training phase of our proposed OLS learning design. For example, the loss function could have a larger weight coefficient for prediction errors that are above a certain threshold (e.g., \cite{mukhoti2020calibrating}). Alternatively, one could directly incorporate the constraint violation into a regularized loss function as in learning-for-OPF work \cite{chen2022learning,liu2022topology,singh2021learning}. For both of these mismatches, it is also possible to consider its worst-case risk by using e.g., the conditional value-at-risk (CVaR) metric; see e.g., \cite{lin2022risk}. These modifications of loss functions would only affect the training phase, and thus still ensure a decentralized OLS in the online implementation.
\end{remark}

\section{Identifiability Analysis}
\label{sec:analysis}

An important aspect of our decentralized OLS is to ensure that using locally available measurements only can differentiate two distinct contingencies. Otherwise, if the local measurements are identical, the ML algorithm may be unable to distinguish which contingency triggered the inputs, potentially causing ambiguity in local load-shedding decisions.
Therefore, this capability to differentiate between contingency scenarios enables more accurate predictions, translating local data into well-informed local decisions. To this end, we investigate the conditions for attaining this identifiability using the simple DC power flow model \cite{stott2009dc}. This analysis will provide useful insights into the realistic scenarios in practical systems. The analysis primarily focuses on contingencies that cause topology changes in the power system. For contingencies such as short circuits and bus faults, their impact on system topology is temporary, as these events typically trigger protective devices to isolate faults automatically \cite{kasikci2018short}. Even so, if any fault in the category leads to the isolation of lines for fault clearance, then the analysis is still applicable.

We start the analysis with the DC model which linearly relates the active power injection $\bbp$ to nodal angle $\bbtheta$ and line flow $\bbf$, as given by
\begin{align} 
\bbp = \bbB\bbtheta,~\textrm{and}~\bbf = \bbK \bbtheta = 
\bbS \bbp 
\end{align}
where $\bbB$ is the B-bus matrix, and matrix $\bbS$ is the so-termed injection shift factor (ISF) matrix.
It allows for a very fast analysis of the post-contingency power flow. This is because the B-bus matrix under any given contingency scenario indexed by $k$ can be updated as $\bbB^{(k)}$, by eliminating the outaged lines in contingency $k$. Assuming the post-contingency system stays connected with no islanding, we can express the difference in line flows as 
\begin{align} 
   \Delta \mathbf{f} & :=  {\mathbf{S}}^{(k)}\mathbf{p} -\mathbf{f}  = \mathbf{K} ([\mathbf{B}^{(k)}]^{-1} - \mathbf{B}^{-1}) \mathbf{p}  \nonumber \\
  & = \mathbf{d}^{(k)}  \mathbf{f}^{(k)} \label{eq:diff_f}
\end{align}
where the vector $\bbf^{(k)}$ contains the pre-contingency flows of the outaged lines under contingency $k$. The latter can be used to fully determine the line flow differences everywhere in the system by using the matrix $\bbd^{(k)}$. This result builds upon the special structure between $[\bbB^{(k)}]^{-1}$ and $(\bbB^{-1})$, based on the well-known matrix inverse lemma \cite{tylavsky1986generalization}. Taking the example
of a single-line contingency $k$ with the line $(i,j)$ in outage, the vector $\bbf^{(k)}$ is a scalar with only the pre-outage flow for line $(i,j)$. 
In this case, matrix $\bbd^{(k)}$ becomes a vector with the length as the total number of lines and is known as the line outage distribution factor (LODF), given by
\begin{align}
    \bbd^{(k)} \coloneqq \frac{({\mathbf{S}}^{(k)} - {\mathbf{S}})\mathbf{p}}{{f}_{ij}} = \frac{{\mathbf{S}_{:i}} - {\mathbf{S}_{:j}}}{1-{\mathbf{S}_{ki}}+{\mathbf{S}_{kj}}} \label{eq:lodf}
\end{align}
where the subscript of matrix $\mathbf{S}$ is used to index its entry with ${:i}$ indicating the $i$-th column. 
Generalizing \eqref{eq:lodf} to a contingency of multiple-line outages is possible but would be more complicated; see e.g., \cite{guler2007generalized,guo2009direct}.

\begin{figure}[t!]
\centering
   \subfloat[]{
      \includegraphics[trim=0cm 0.5cm 0cm 0.5cm,clip=true, width=0.18\textwidth]{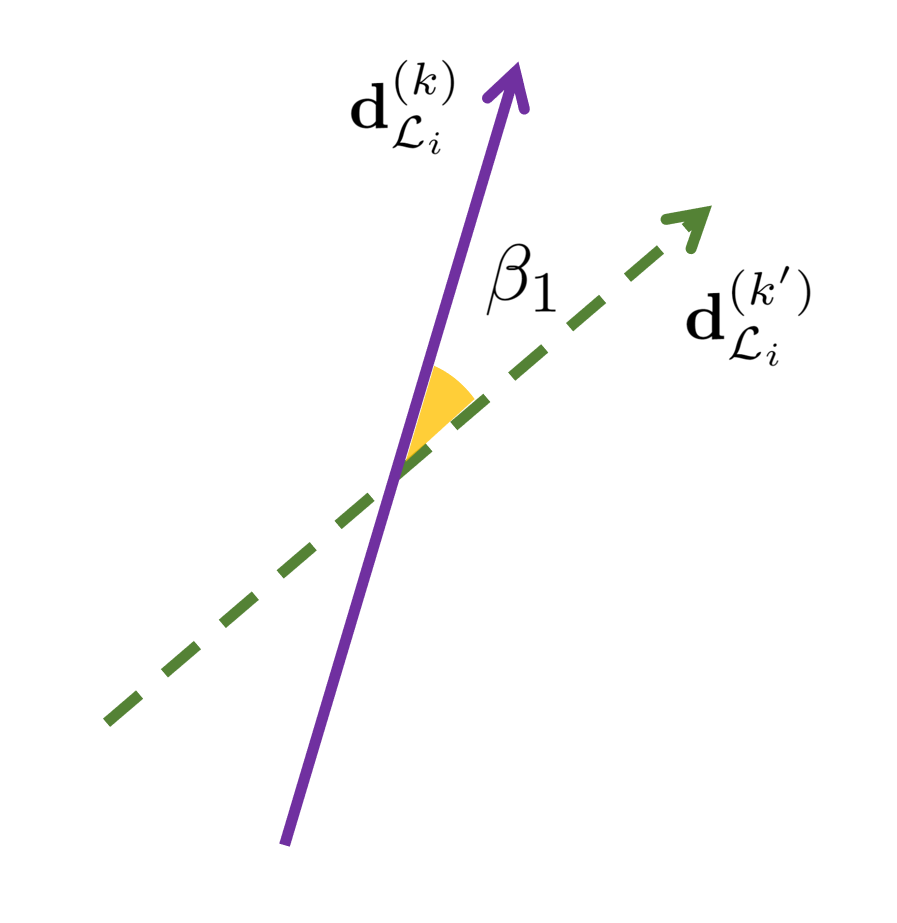}}
\hspace{0.2cm}
   \subfloat[]{
      \includegraphics[trim=0cm 0.5cm 0cm 0.5cm,clip=true, width=0.26\textwidth]{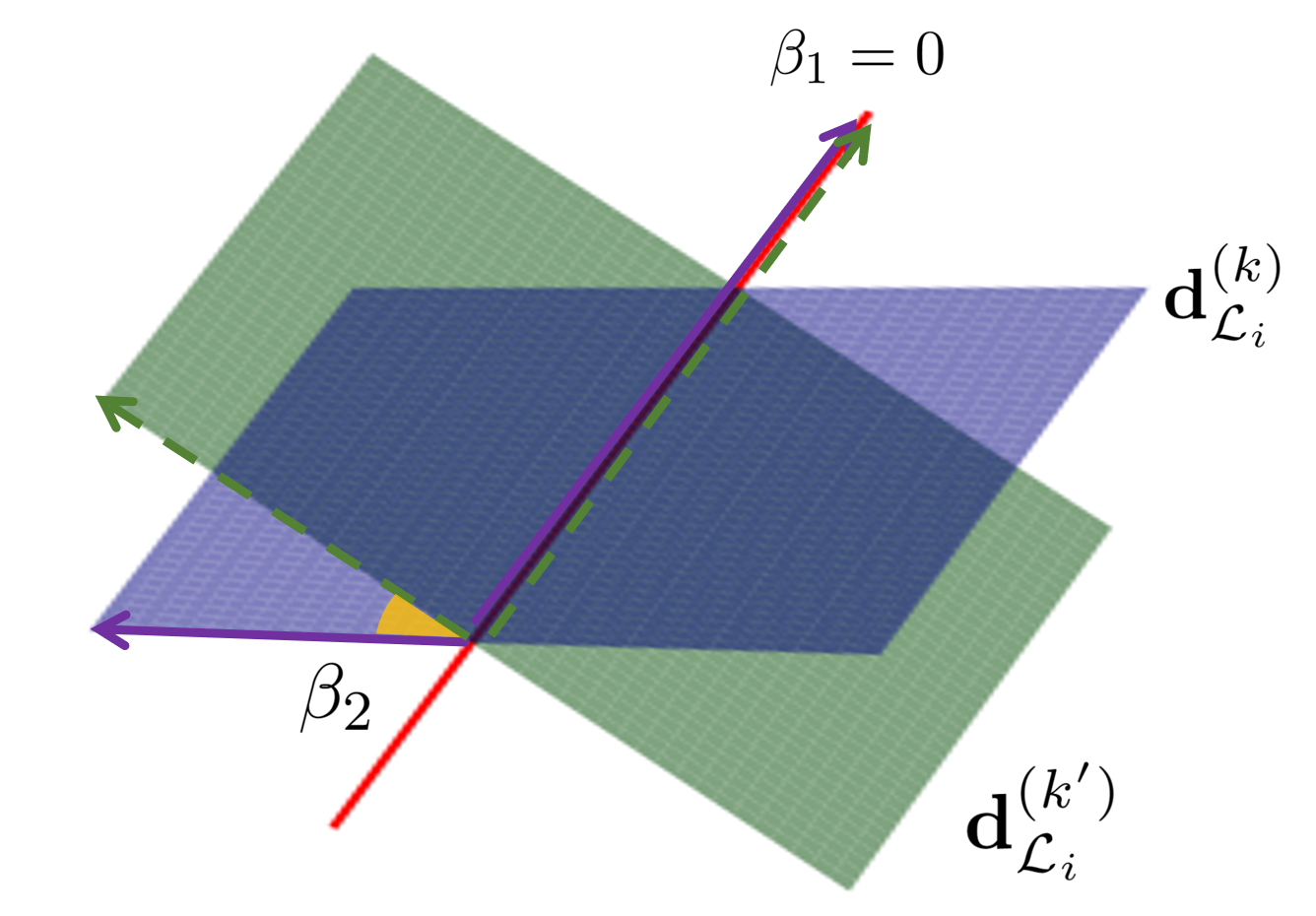}}\\
\caption{The principal angles between the two subspaces spanned by ${\mathbf{d}}_{{\cal{L}}_i}^{(k)}$ and  ${\mathbf{d}}_{{\cal{L}}_i}^{(k')}$, for the case of (a) 1-D subspaces with $\beta_1>0$; and (b) 2-D subspaces with $\beta_1=0$ and $\beta_2>0$.}
\label{fig:subspace}
\end{figure}

We can utilize the matrix $\bbd^{(k)}$ to determine whether there is sufficient identifiability of different contingencies. Note that only limited measurements are available to each load center, thus we consider a subset of lines denoted by ${{\cal{L}}}_{i} \subset {\cal L} $, as the local incident lines to bus $i$, for this analysis. 
This way, the change of power flow pattern in the local region of bus $i$ is given by:
\begin{align} 
 \Delta \mathbf{f}_{\mathcal{L}_i} = {\mathbf{d}}_{{\cal{L}}_i}^{(k)} \mathbf{f}^{(k)} = \left[\{\mathbf{d}_{\ell:}^{(k)}\}_{\ell  \in {\cal{L}}_i}
 \right]
 \mathbf{f}^{(k)} \label{eq:busi}
\end{align}
by selecting only the lines $\ell\in {\cal{L}}_i$ to form the local flow vector $ \Delta \mathbf{f}_{\mathcal{L}_i}$ and submatrix ${\mathbf{d}}_{{\cal{L}}_i}^{(k)}$ of sensitivity factors. Thus, the key for bus $i$ to be able to differentiate two contingency scenarios, namely $k$ and $k'$, requires a sufficient level of dissimilarity in some sense, between the two corresponding matrices ${\mathbf{d}}_{{\cal{L}}_i}^{(k)}$ and  ${\mathbf{d}}_{{\cal{L}}_i}^{(k')}$.  We will formalize this condition.

First, consider the simple case of single-line outage scenarios, where the two matrices are essentially vectors of the same length. 
In this case, as long as the vector ${\mathbf{d}}^{(k)}_{{\cal{L}}_i}$ is not in parallel with vector $\mathbf{d}^{(k')}_{{\cal L}_i}$, their corresponding subspaces of local flow vectors will not overlap. As illustrated in Fig.~\ref{fig:subspace}(a), this is equivalent to requiring the acute angle (formally defined as the principal angle soon) between these two subspaces to be \textit{non-zero}, which is satisfied by having 
\begin{align} 
\frac{\left|\left[{\mathbf{d}}^{(k)}_{{\cal L}_i}\right]^{\mathsf T}{\mathbf{d}}^{(k')}_{{\cal L}_i}\right|}{\left\| {\mathbf{d}}^{(k)}_{{\cal L}_i} \right\|\cdot \left\|{\mathbf{d}}^{(k')}_{{\cal L}_i}\right\|} \neq 1.
\label{eq:cos}
\end{align}
This condition guarantees that the change of local flows in bus $i$ under the two contingencies $k$ and $k'$ will not be the same. As a result, the local flow measurements alone are sufficient for distinguishing the specific contingency, and accordingly, for determining the corresponding OLS corrective actions. 

To consider multiple-line outage contingencies, we can generalize \eqref{eq:cos} to the matrix case.
Given two matrices $\mathbf{d}^{(k)}_{{\cal L}_i}$ and $\mathbf{d}^{(k')}_{{\cal L}_i}$, the identifiability depends on the principal angles between the corresponding subspaces \cite{miao1992principal}. To this end,  let us use the QR decomposition to obtain the orthonormal bases of the two subspaces, as given by
\begin{align}
   \mathbf{d}^{(k)}_{{\cal L}_i}  = \mathbf{Q}^{(k)} \mathbf{R}^{(k)},~\textrm{and}~
    \mathbf{d}^{(k')}_{{\cal L}_i}  = \mathbf{Q}^{(k')} \mathbf{R}^{(k')}
\end{align}
where $\mathbf{Q}^{(k)}$ or $\mathbf{Q}^{(k')}$ is the orthornormal basis, while $\mathbf{R}^{(k)}$ or $\mathbf{R}^{(k')}$ is an upper triangular matrix. Clearly, the dimension of each subspace can be as large as the number of outaged lines in the contingency. We will use $|k|$ and $|k'|$ to denote the dimensions of these two subspaces, respectively.

We can further characterize the separation between the two subspaces via the singular value decomposition (SVD), as \cite{golub1995canonical}
\begin{align} 
  \left[\mathbf{Q}^{(k)}\right]^{\mathsf T}\mathbf{Q}^{(k')} = \mathbf{U} \boldsymbol{\Sigma} \mathbf{V}^{\mathsf T} \label{eq:svd}
\end{align}
with all the singular values stored in the diagonal matrix $\boldsymbol{\Sigma}$. Here, the number of singular values $s = \min(|k|,|k'|)$, which is the minimum of the two subspaces' dimensions. The ordered singular values $\sigma_1\geq \sigma_2\geq \cdots\geq\sigma_{s}\geq 0$
allow us to find the vector of principal angles in a \textit{non-decreasing order}, as denoted by
\begin{align}
 \bbbeta:=[\beta_1, \cdots, \beta_{s}]^{\mathsf T}=[\arccos(\sigma_1), 
 \cdots, \arccos(\sigma_{s})]^{\mathsf T}. \label{eq:sigma}
\end{align}
This vector $\bbbeta$ can fully characterize the angular separation between the two subspaces of interest. 
For the aforementioned single-line outages with $|k|=|k'|=1$, the only singular value $\sigma_1$ exactly equals the ratio term given in \eqref{eq:cos}. As shown in Fig.~\ref{fig:subspace}(a),  the orthonormal bases of vectors $\mathbf{d}^{(k)}_{{\cal L}_i}$ and $\mathbf{d}^{(k')}_{{\cal L}_i}$ are essentially formed by normalizing each of them to have unit norm. Hence, if \eqref{eq:cos} holds, we have $\sigma_1 \neq 1$, and  accordingly, the only principal angle $\beta_1 = \arccos(\sigma_1)$ is nonzero. Interestingly, extending this result to higher-dimensional subspaces requires having at least one non-zero principal angle in $\bbbeta$, or equivalently, at least one $\sigma_\ell$ in \eqref{eq:sigma} to be different from $1$.  Fig.~\ref{fig:subspace}(b) shows the example of two 2-D subspaces, where the smaller angle $\beta_1=0$ but the larger one $\beta_2\neq 0$ can ensure a sufficient level of subspace separation. This result directly follows from \cite{golub1995canonical} and allows to formalize the identifiability for different contingencies, as stated below.

\begin{figure*}[t!]
\centering
   \subfloat[]{
      \includegraphics[trim=0cm 0cm 0cm 0cm,clip=true, width=0.32\textwidth]{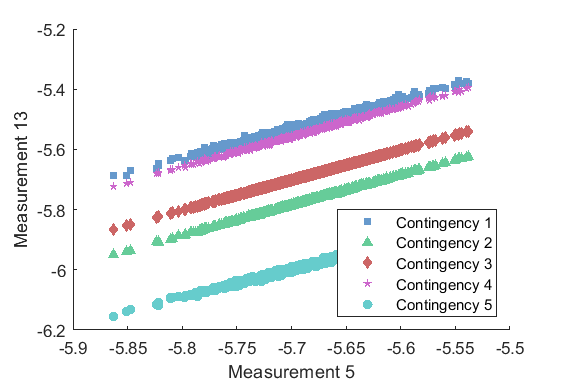}}
\hspace{\fill}
   \subfloat[]{
      \includegraphics[trim=0cm 0cm 0cm 0cm,clip=true, width=0.32\textwidth]{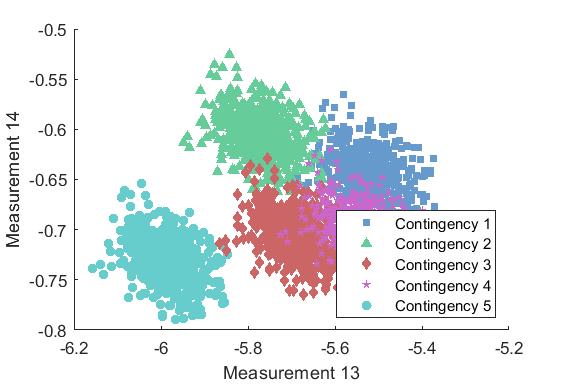}}
\hspace{\fill}
   \subfloat[]{
      \includegraphics[trim=0cm 0cm 0cm 0cm,clip=true, width=0.32\textwidth]{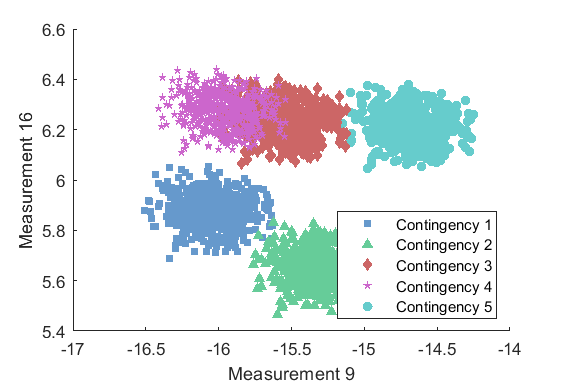}}\\
\caption{{The scatter plots of the local measurements at Bus 34 for the 5 contingencies considered for the IEEE 118-bus system based on a combination of only two measurements indexed by: (a) 5 and 13; (b) 13 and 14; and (c) 9 and 16.}}
\vspace*{-3mm}
\label{fig:classification}
\end{figure*}

\begin{proposition}[Identifiability for two contingencies] \label{eq:prop1}
To guarantee that the local line flow data at bus $i$ are sufficient for differentiating the two contingencies $k$ and $k'$, we need 
\begin{align}
\exists \, \ell \in \{1,\ldots, s\}, \textrm{~such~that~} \sigma_{\ell} \neq 1  \label{eq:cond}
\end{align}
for at least one singular value found by \eqref{eq:svd}. This ensures a sufficient separation between the two subspaces spanned by $\mathbf{d}^{(k)}_{{\cal L}_i}$ and $\mathbf{d}^{(k')}_{{\cal L}_i}$ with at least one of the principal angles $\beta_\ell \neq 0$. 
\end{proposition} 

\begin{corollary}[Identifiability for a contingency set] \label{eq:coro}
For a given contingency set $\ccalK$, the local data can yield unique outputs for each scenario in $\ccalK$ as long as every two contingency scenarios $k,k'\in\ccalK$ can be differentiated per Proposition~\ref{eq:prop1}.
\end{corollary} 
 
Proposition~\ref{eq:prop1} and Corollary \ref{eq:coro} have provided the analytical conditions for using only local data to differentiate the possible contingencies. Specifically, \eqref{eq:cond} follows directly from the classical results on the angular separation between two linear subspaces in \cite{miao1992principal,golub1995canonical}, while Corollary \ref{eq:coro} naturally extends it to the case of multiple subspaces. Note that Proposition \ref{eq:prop1}  is also equivalent to
\begin{align}
    \min \, \{\sigma_1, \sigma_2, \cdots, \sigma_{s} \} <1 ,
\end{align}
which ensures the existence of at least one non-zero principal angle. This way, neither of the two subspaces can be entirely contained by the other one, and thus it is possible to distinguish them using a sufficient sampling of the two subspaces.

These analytical conditions can support the feasibility of our scalable OLS learning concept, by ensuring that local OLS decision rules can potentially infer the specific contingency and accordingly determine the best OLS actions. Of course, it can be sometimes challenging to satisfy these conditions. With only one measurement at bus $i$, it is almost impossible to differentiate any two contingencies. This is because in this case, both $\mathbf{d}^{(k)}_{{\cal L}_i}$ and $\mathbf{d}^{(k')}_{{\cal L}_i}$ in \eqref{eq:cos} would become scalars, and thus the ratio term therein has to be 1 which yields $\beta_1=0$. 
Having multiple measurements at bus $i$ can, in  most of the circumstances, lead to a non-zero $\beta_1$ for two vectors. The only circumstance having $\beta_1=0$  is when the two subspaces fully coincide with each other, which could be more unlikely when the number of measurements increases. In this sense, it would be ``easier'' to satisfy the condition \eqref{eq:cond} with multiple line-outages, because with an increasing dimension of subspaces, we only need one of the principal angles to be non-zero. In general, the more diverse measurements bus $i$ could access, the more likely the identifiability condition would hold to support our scalable OLS design. Note that the number of contingencies for a given networked system can be extremely high, especially when we consider multiple simultaneous component failures. However, in actual power system operations, ISOs typically restrict their analysis to N-1 and N-2 contingencies \cite{chen2008risk,isone2023}, with N-3 considered only in rare cases. In addition, ISOs normally have a prioritized list of contingencies that they use to ensure reliable grid operations. These contingency scenarios are part of their contingency analysis and planning procedures \cite{chen2014contingency}, which help them maintain system stability and prepare for possible disruptions. Contingency prioritization \cite{fliscounakis2013contingency} also takes into account events with specific risks, such as extreme weather events, and seasonal load fluctuations. Based on grid conditions, congestion, or maintenance plans, the ISOs will also adjust their priorities in real time. Without loss of generality, this work assumes that a subset of prioritized contingencies have been defined by system operators, which will be the primary focus of the learning process. Then again, since the learning process algorithm is mainly conducted offline using massive datasets, it can effectively cover most common system contingencies (e.g., N-1, N-2) to support real-time decision-making.

\subsection{Numerical Example}
\label{sec:class}

We present one specific numerical example to corroborate that local data can be sufficient for distinguishing different contingency scenarios using the IEEE 118-bus system. This system consists of 186 lines and 19 conventional generators, with the remaining 99 buses as load centers. Specifically, we want to demonstrate that local data available at the load center of interest, bus 34, are sufficient for training a very accurate classifier that can distinguish all the scenarios in a contingency list. 
We have selected 5 contingency scenarios, including 3 single-line outage scenarios and 2 double-line outage ones. The outaged lines are chosen by prioritizing those with the highest nominal line flows, and thus the resultant contingencies tend to have a very large impact on the overall system safety and would require a high level of load shedding. To train this classifier, the input features include all the incident line flows (active/reactive powers) during both pre- and post-contingency operations provided by the power flow simulators. For each contingency scenario, we randomly generate 1000 samples by adding some small perturbations to the steady-state loads. A simple feed-forward NN is trained using $70\%$ of the total samples for this classification task. It is shown to exhibit an outstanding prediction performance when tested on the remaining $30\%$ of samples, achieving a perfect accuracy of $100\%$. 
To illustrate this result, Fig.~\ref{fig:classification} plots the data samples for those 5 contingencies in a 2-D space,  based on selected measurement pairs. We can observe that even with just two channels of local data, each of the contingencies manifests themselves with differentiable patterns of clusters or shapes in these 2-D distribution plots. These unique patterns have well supported our analysis that local data could provide sufficient identifiability for different contingency scenarios. Notably, this capability is truly important for our decentralized OLS  design. It can ensure that even if trained with local data only, the resultant OLS decision rules are able to predict the specific contingency, possibly elsewhere in the system, and further effectively predict the globally optimal OLS actions. The ensuing section will present the comprehensive validation of the proposed scalable OLS design in terms of directly predicting OLS-related decisions.

\section{Simulation Results}
\label{sec:sr}

This section presents the numerical results for our proposed AC-OLS learning approach using the IEEE 118-bus system and a larger 2000-bus system representing the Texas system \cite{birchfield2016grid}. The AC-OLS problem \eqref{eq:OLS} is implemented in the  MATPOWER simulator \cite{zimmerman2010matpower}
by its primal-dual interior point method solver, while the post-contingency measurements are obtained by MATPOWER's power flow solver. The load shedding and reserve cost coefficients have been specified by modifying the quadratic cost functions in the original test cases to satisfy \eqref{eq:con_1}-\eqref{eq:con_2}. Basically, the marginal cost of load shedding has been set to be higher than that of reserve flexibility, to ensure the latter is fully utilized before shedding the load. 
Note that the AC-OLS solver can also return the multiplier $\alpha_i$ for the active power balance equation \eqref{eq:OLS_h}, which will be used as the output label for the NN training. The feedforward NN for our OLS design has been implemented using the MATLAB\textsuperscript{\textregistered} deep learning toolbox with the Bayesian regularization. 
We have adopted a standard NN setup with 3 hidden layers. For the 2000-bus system, a 3-layer setup turns out to be adequate, even though each load center has dozens of input features (local measurement channels). 
All the simulations have been conducted on a regular laptop with Intel\textsuperscript{\textregistered} CPU @ 2.60 GHz and 16 GB of RAM.

\subsection{118-bus System Tests}

We have used the IEEE 118-bus system \cite{118bus} for validating the scalable OLS design in addition to classifying the contingencies in Section~\ref{sec:class}.
To solve the AC-OLS problem \eqref{eq:OLS}, we consider a total of 6 contingency scenarios. The majority of them are single-line or double-line outages, with one scenario having 3 outaged lines. 
We have chosen the outaged lines by prioritizing those with the highest nominal line flows. This is because the resultant contingencies would have a significant impact on the overall safety of the system, necessitating a higher level of load shedding. For each contingency scenario, we generate a total of 2000 samples by randomly perturbing the pre-outage load profile.  Specifically, the load demand at each bus has been changed randomly within the range of $[0.95,1.05]$ times its nominal value. To train the OLS decision rules, the pre- and post-contingency local data serve as the input features at each load center, including local demand and incident line flows (real/reactive). The dimension of these local inputs ranges from 5 to 49, depending on the specific bus location. The output label is the corresponding multiplier $\alpha_{i}$ at bus $i$, as returned by the AC-OLS solver. A feedforward NN with 3 hidden layers (with 40, 30, and 20 neurons) has been trained using $70\%$ of the total samples, with the remaining $30\%$ reserved for testing. It turns out this standard NN setup can attain very high prediction performance later on. 

\begin{figure}[t!]
\centering
\includegraphics[trim=0cm 0.1cm 0cm 0cm,clip=true,totalheight=0.20\textheight]{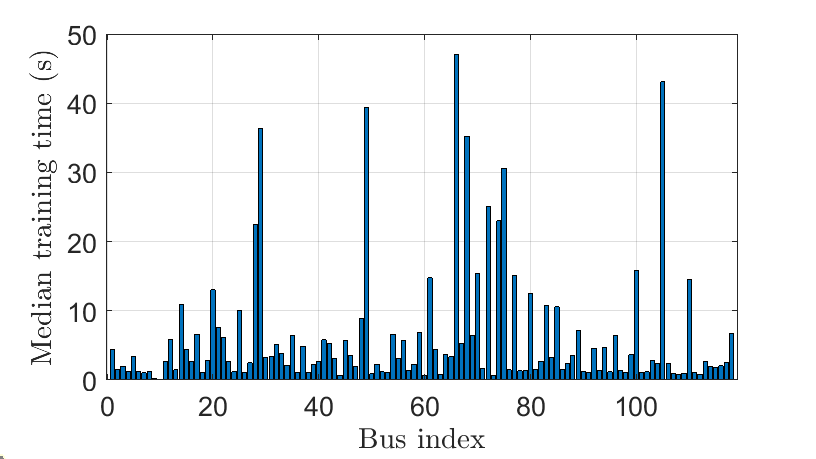}
\caption{The median value of training time (in seconds) for each bus in the 118-bus system.}
\vspace*{-2mm}
\label{fig:118bus_4}
\end{figure}

To first demonstrate the computational efficiency, we have conducted multiple runs of NN training for every bus, and plot the median training time in Fig.~\ref{fig:118bus_4}.
Clearly, the majority of buses enjoy extremely fast training time within a few seconds, with the maximum training time being less than 1 minute. This high computation speed is achieved thanks to the proposed scalable OLS design, under which the input features only depend on the local measurements available at each bus and thus are much less than the total number of measurements throughout the overall system.

\begin{figure}[t!]
\centering
   \subfloat[]{%
      \includegraphics[trim=0cm 0cm 0cm 0cm,clip=true, width=0.45\textwidth]{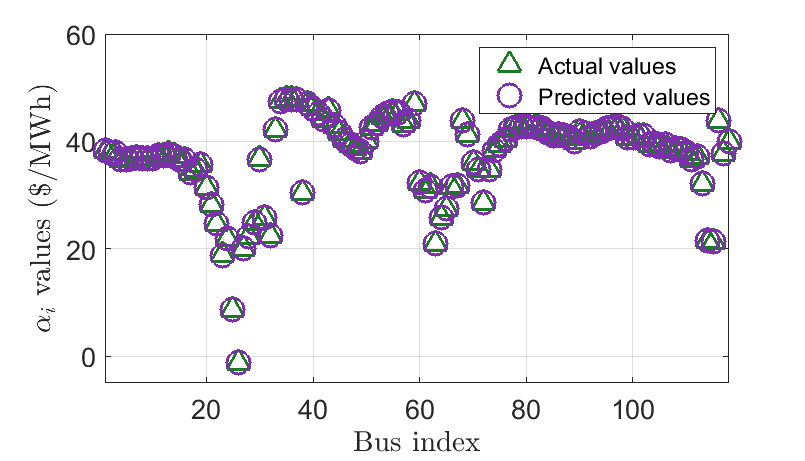} \label{fig:ols_118_a}%
   }\hspace{\fill}
   \subfloat[]{%
      \includegraphics[trim=0cm 0cm 0cm 0cm,clip=true, width=0.45\textwidth]{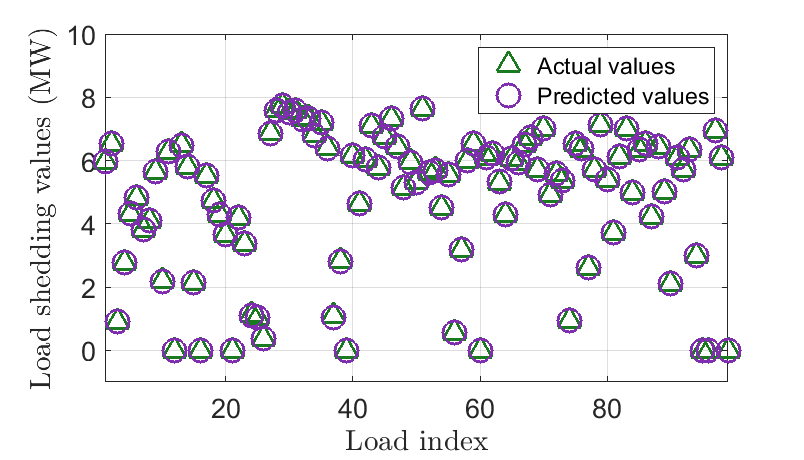} \label{fig:ols_118_b}%
   }
\caption{Prediction on (a) Lagrange multipliers; and (b) load shedding; across different buses in the IEEE 118-bus system.}
\label{fig:118_prediction}
\end{figure}
\begin{figure}[t!]
\centering
\includegraphics[trim=0cm 0cm 0cm 0cm,clip=true,totalheight=0.20\textheight]{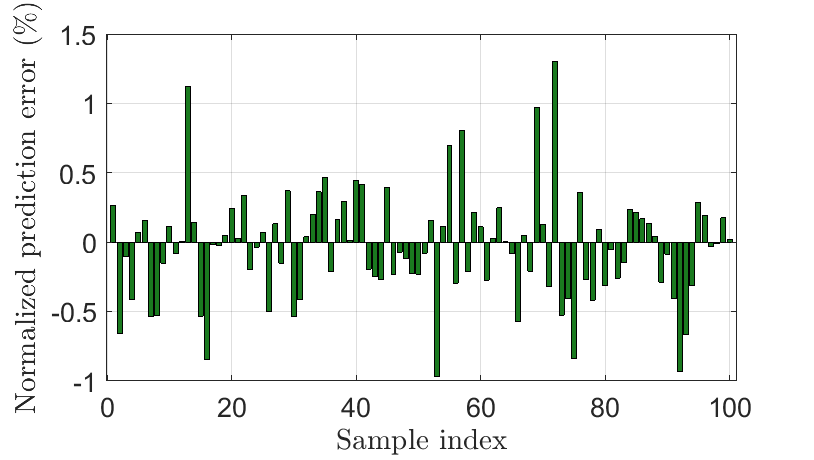}
\caption{Normalized prediction error of Lagrange multiplier at Bus 43 under randomly selected testing samples.}
\vspace*{-2mm}
\label{fig:118bus_3}
\end{figure}

We further demonstrate the effectiveness of our OLS design in predicting both the multiplier $\alpha_i$ and the OLS decision $\hat{p}^s_i$. Recall that the proposed OLS learning utilizes the KKT optimality condition in  \eqref{eq:op_1}, where the value of $\alpha_i$ can fully determine $\hat{p}^s_i$. Fig.~\ref{fig:118_prediction} compares the actual value with the predicted one using one representative instance of the testing sample. Note that while the $\alpha_i$  prediction can be performed for every bus, the prediction of load shedding amount $\hat{p}^s_i$ is only considered for the load bus only, and thus the x-axis of Fig.~\ref{fig:118_prediction}(b) only lists the 99 loads in the 118-bus system.  Fig.~\ref{fig:118_prediction}(a) shows that the multipliers do have some strong locational dependence. While at most buses $\alpha_i$ is around 40\$/MWh,  some of the buses have very different values that can go as low as 0\$/MWh.  This is because the system congestion pattern could make certain buses much more preferred than the other for load shedding, and those preferable buses would have relatively lower $\alpha_i$ values. We can observe that even with this high variability of $\alpha_i$ values,  our OLS design can attain a very high prediction accuracy at all the buses. The average prediction error is  0.2716\$/MWh out of the average multiplier value of 36.8865\$/MWh, with an error percentage of $0.74\%$. Thanks to this excellent performance in predicting $\alpha_i$, the OLS decisions $p_i^s$ can be also well predicted as shown in Fig.~\ref{fig:118_prediction}(b). Similarly, there is also a high level of locational difference, as certain loads could more critically reduce the system-wide congestion during the post-contingency operations than the others. But still, the average prediction error for $\hat{p}^s_i$ is very low at 0.0621MW, with a maximum error of 0.48MW. Note that all these prediction results have been obtained from local data only available at each bus, and thus they have verified the effectiveness of our scalable OLS design.

\begin{figure}[t!]
\centering
\includegraphics[trim=0cm 0cm 0cm 0cm,clip=true,totalheight=0.18\textheight]{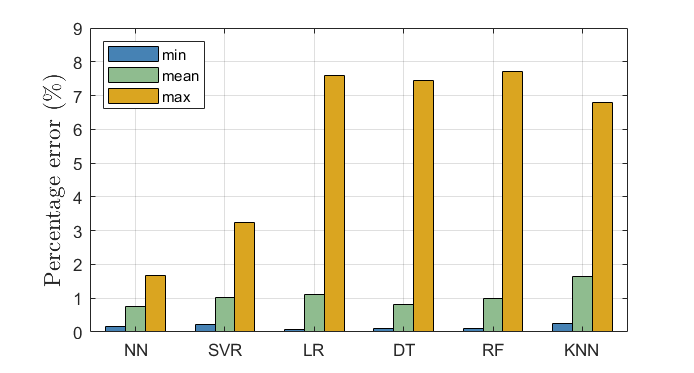}
\caption{{Comparison of percentage errors over all buses in the 118-bus system under selected machine learning algorithms.}}
\vspace*{-2mm}
\label{fig:comparison}
\end{figure}

\begin{figure*}[t!]
\centering
\includegraphics[trim= 2cm 0cm 2cm 0cm,clip=true,totalheight=0.22\textheight]{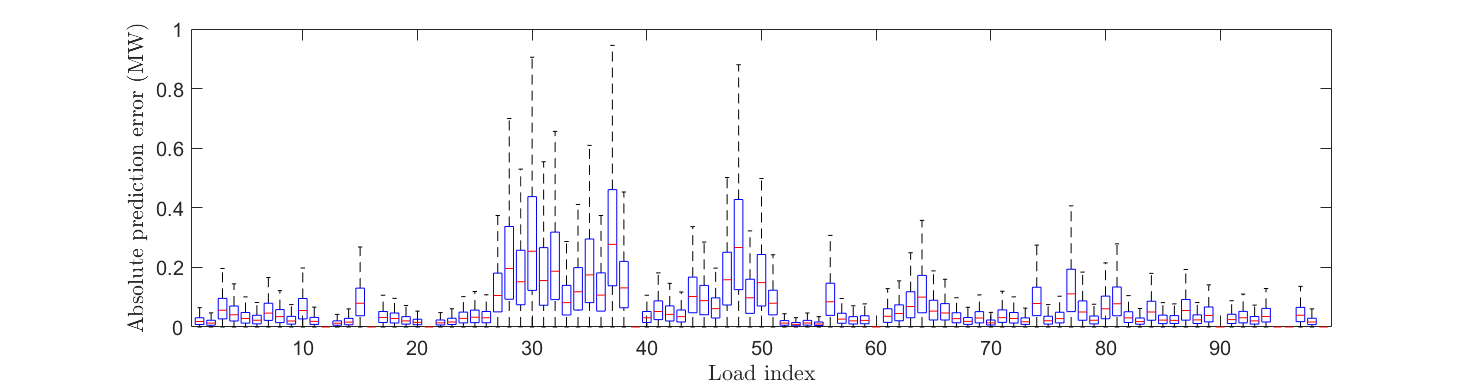}
\caption{Absolute value of prediction errors for $p_i^s$ in box plot at every load bus in the 118-bus system.}
\vspace*{-2mm}
\label{fig:118bus_5}
\end{figure*}

Moreover, Fig.~\ref{fig:118bus_3} shows the prediction performance across all samples by plotting the normalized prediction error attained at bus 43. This load center is a very representative one in the system which has consistently non-zero  $\hat{p}^s_i$ across all samples. Since the $\alpha_i$ prediction is very critical for our overall design, we only plot the error percentage as normalized by the actual value for the task of predicting $\alpha_i$ only.  Only 100 samples from the testing dataset have been selected here due to the size limit. But for all testing samples have very minimal prediction error at a similar level to those shown in  Fig.~\ref{fig:118bus_3}. This prediction error tends to fall within the range of $[-1\%,~1\%]$, and its absolute value has an average of only $0.29\%$. This result again shows a very high accuracy in predicting $\alpha_i$ across all contingency scenarios and operating conditions.

Finally, we present a comprehensive analysis of the OLS decision prediction performance in Fig.~\ref{fig:118bus_5}, which plots the error statistics of predicting $\hat{p}^{s}_{i}$ for each load bus. Specifically, a box plot is used to mark the median, first, and third quartiles, as well as the minimum and maximum, of the absolute values of prediction errors across all testing samples. The median values range from almost 0MW to 0.28MW, with an average of 0.054MW across all load buses. For the first and third quartiles, they respectively have an average of 0.026MW and 0.092MW across all load buses. Based on an average load shedding amount of 4.81MW, this prediction has a very high accuracy for a majority of the load buses.  Notably, a few of the load buses exhibit relatively higher errors than the average case, likely due to having a larger variability of the OLS decisions. 
Overall, the prediction accuracy for $\hat{p}^{s}_{i}$ has been pretty consistently maintained across all load buses, demonstrating the effectiveness of our proposed OLS learning on the 118-bus system. To showcase that the NN model provides an accurate prediction performance, we further include some comparison results from other regression models. Specifically, we have considered support vector regression (SVR), linear regression (LR), decision tree (DT), random forest (RF), and k-nearest neighbors (KNN), to compare with the benchmark NN model, in terms of prediction performance. The results are shown in Fig. \ref{fig:comparison}. Across all buses, the NN model achieves an average prediction error of $0.74\%$, while other ML algorithms have slightly higher prediction errors which are $1.01\%$, $1.10\%$, $0.80\%$, $0.99\%$, $1.64\%$, respectively. In addition, the NN model maintains minimal variance in prediction errors compared to the other models, which makes it well-suited under the decentralized learning setup. Next, we will assess the performance of the scalable OLS algorithm using a more realistic 2000-bus test case.

\subsection{2000-bus System Tests}

The ACTIVSg2000 test case \cite{birchfield2016grid} is a synthetic model, constructed using public data and statistical analyses of actual power systems.
Developed to represent the geographical coverage of Texas, this system consists of 2000 buses and 3206 transmission lines. We once again consider a total of 6 contingency scenarios, with most of them being single-line or double-line outages, while one scenario includes 3 outaged lines. Similarly, the outaged lines are chosen by prioritizing those with the highest nominal line flows. To maintain the sample size, the samples are generated by focusing on the top 400 most heavily loaded buses by randomly perturbing the load demands therein within $[0.97,1.03]$ of the nominal values. The training and testing of local OLS policies follow similarly from the 118-bus tests. The dimension of local input features at each bus increases to as high as 69, due to the presence of a more complicated network topology in this system. Thus, we use a feedforward NN with 3 hidden layers, with each of them having 50, 30, and 20 neurons, respectively. With the presence of thousands of load centers, we have selected a subset of them to showcase the test performance. Specifically, the load centers exhibiting the largest variability of the multiplier $\alpha_i$ have been selected, as they pose higher challenges in the OLS prediction as compared to load centers with pretty constant multipliers. The OLS training for the selected load centers takes 116.57s  on average, with the fastest and slowest being 28.37s and 669.42s, respectively. This training time is slightly higher than that in the 118-bus system, mainly due to increased dimensionality. However, this increase of time is reasonable compared to the system size, corroborating the scalability of our proposed OLS design.

\begin{figure}[t!]
\centering
\includegraphics[trim=0cm 0cm 0cm 0cm,clip=true,totalheight=0.17\textheight]{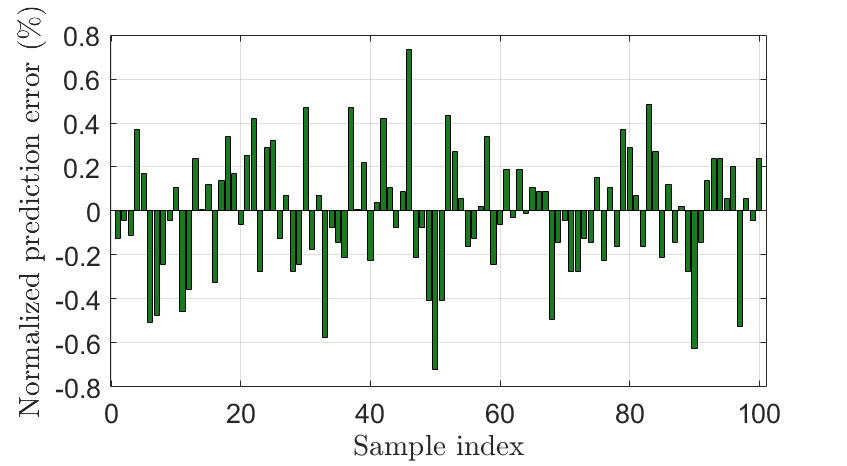}
\caption{Normalized prediction error of Lagrange multiplier at Bus 68 under randomly selected testing samples.}
\vspace*{-2mm}
\label{fig:2000bus_1}
\end{figure}

\begin{figure}[t!]
\centering
\includegraphics[trim=0cm 0cm 0cm 0cm,clip=true,totalheight=0.17\textheight]{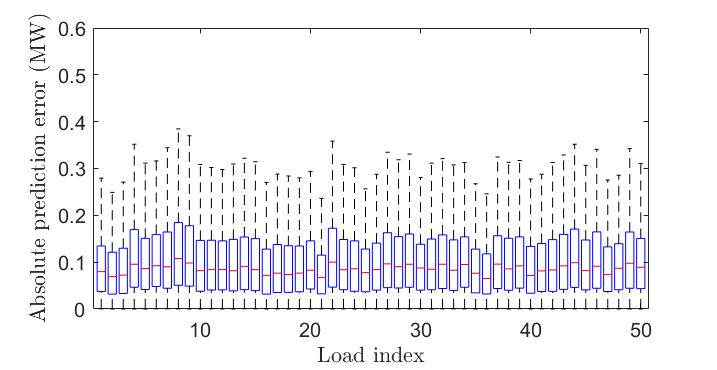}
\caption{Absolute value of prediction errors for $p_i^s$ in box plot at selected load buses in the 2000-bus system.}
\vspace*{-2mm}
\label{fig:2000bus_2}
\end{figure}

We first demonstrate the prediction performance across samples at bus 68, which features a comparably small set of 9 local inputs only. This particular load center is representative, as characterized by consistently non-zero, highly variable,  load shedding amount. Similar to Fig.~\ref{fig:118bus_3} for the 118-bus system, Fig.~\ref{fig:2000bus_1} here plots the error percentage normalized by the actual value in predicting $\alpha_i$, based on a total of 100 samples randomly chosen from the testing dataset. The prediction performance for the majority of the testing samples exhibits a similar error level to those in Fig.~\ref{fig:2000bus_1}. We observe that the prediction errors lie within the range of $[-0.8\%,~0.8\%]$, with the absolute error percentages averaging at only $0.22\%$. This prediction error for the 2000-bus test case is slightly smaller than that in the 118-bus case. These results demonstrate exceptionally high accuracy in predicting the multiplier $\alpha_i$ and confirm that our proposed OLS design can well maintain the scalability with respect to system size.

Furthermore, we present a comprehensive validation of the prediction performance at multiple load buses. Specifically, we select the top 50 load buses with the highest variability in their local multiplier. Accordingly, the variability of load shedding amounts at these selected buses are also among the highest due to the relation between these two variables in \eqref{eq:op_1}. 
Fig.~\ref{fig:2000bus_2} shows the box plot of OLS prediction error for $\hat{p}^{s}_{i}$, with the median, first and third quartiles, and the minimum and maximum of the absolute prediction errors. The median prediction errors range from 0.065MW to 0.108MW, with an average of 0.085MW across these load buses. In addition, the first and third quartiles have averages of 0.041MW and 0.148MW, respectively. Using the median error statistic, the OLS prediction across these 50 load centers has an extremely high accuracy of around $0.3\%$  based on an average load shedding amount of 28.17MW. Overall, the consistency in prediction accuracy has been validated, demonstrating the effectiveness of our proposed OLS learning approach for the large 2000-bus system.

\section{Conclusion} \label{sec:con}
In this paper, we developed a decentralized framework for performing real-time optimal load shedding to recover power balance and prevent cascading failures under emergency events. We start by formulating an AC power flow-based optimal load shedding (AC-OLS) problem. By solving AC-OLS repeatedly across diverse contingency scenarios, we propose a learning framework that maps from the local measurements of each load center to the corresponding Lagrange multiplier that exhibits locality property. The OLS decisions can be accurately determined by mapping from the predicted Lagrange multiplier using optimality conditions. The main contribution of this work lies in the scalable decentralized OLS design which enables each load center to react promptly to emergencies without reliance on central supervision. Numerical studies on the IEEE 118-bus and Texas 2000-bus systems have demonstrated the effectiveness of our proposed learning approach in accurately predicting the OLS solutions. Specifically, the off-line training time is shown to scale well with the system size thanks to the decentralized design. Meanwhile, the prediction accuracy is very high on average (at a 0.22\% of absolute error for the 2000-bus system). By increasing the number of training samples and selecting the candidate contingencies, the potential impact on prediction accuracy could be further mitigated. For future work, we plan to investigate topology-aware OLS learning approaches and work on improving the dependability of OLS predictions through robust risk-aware design.

\bibliography{bibliography.bib}

\bibliographystyle{IEEEtran}

\itemsep2pt

\end{document}